\documentclass[conference]{IEEEtran}
\IEEEoverridecommandlockouts
\usepackage{cite}
\usepackage{amsmath,amssymb,amsfonts}
\usepackage{algorithmic}
\usepackage{graphicx}
\usepackage{textcomp}
\usepackage{xcolor}
\usepackage{hyperref}
\usepackage{subcaption}
\usepackage{multirow}

\def\BibTeX{{\rm B\kern-.05em{\sc i\kern-.025em b}\kern-.08em
    T\kern-.1667em\lower.7ex\hbox{E}\kern-.125emX}}
\begin{document}

\title{HDC-MiniROCKET: Explicit Time Encoding in Time Series Classification with Hyperdimensional Computing
\thanks{$^*$all authors are with Chemnitz University of Technology, Germany
        {\tt\small \emph{firstname}.\emph{lastname}@etit.tu-chemnitz.de}}%
}

\author{Kenny Schlegel, Peer Neubert and Peter Protzel$^{*}$}
        
\maketitle

\begin{abstract}
Classification of time series data is an important task for many application domains.
One of the best existing methods for this task, in terms of accuracy and computation time, is MiniROCKET.
In this work, we extend this approach to provide better global temporal encodings using hyperdimensional computing (HDC) mechanisms.
HDC (also known as Vector Symbolic Architectures, VSA) is a general method to explicitly represent and process information in high-dimensional vectors.
It has previously been used successfully in combination with deep neural networks and other signal processing algorithms. 
We argue that the internal high-dimensional representation of MiniROCKET is well suited to be complemented by the algebra of HDC.
This leads to a more general formulation, HDC-MiniROCKET, where the original algorithm is only a special case.
We will discuss and demonstrate that HDC-MiniROCKET can systematically overcome catastrophic failures of MiniROCKET on simple synthetic datasets.
These results are confirmed by experiments on the 128 datasets from the UCR time series classification benchmark.
The extension with HDC can achieve considerably better results on datasets with high temporal dependence without increasing the computational effort for inference.
\end{abstract}

\section{Introduction}
Time series classification has a wide range of applications in robotics, autonomous driving, medical diagnostic, in the financial sector, and so on. 
As elaborated in \cite{Bagnall2017}, classification of time series differs from traditional classification problems because the attributes are ordered.
Hence, it is crucial to create discriminative and meaningful features with respect to the specific order in time. 
Over the past years, various methods for classification of univariate and multivariate time series have been proposed (for instance, \cite{Schafer2015, Schafer2017,Bostrom2015,Shifaz2020,Middlehurst2021,IsmailFawaz2020,Dempster2020,Dempster2021,Lines2018,Schafer2017b}).
Often, a high accuracy of a method comes at the cost of a high computational effort.
A very noticeable exception is MiniROCKET~\cite{Dempster2021} which superseded the earlier ROCKET~\cite{Dempster2020} and  achieves state-of-the-art accuracy at very low computational complexity.
Similar to a convolutional neural network (CNN) layer, MiniROCKET applies a set of parallel convolutions to the input signal.
To achieve a low runtime, two important design decisions of MiniROCKET are (1) the usage of convolution filters of small size and (2) accumulation of filter responses over time based on the \textit{Proportion of Positive Values (PPV)}, which is a special kind of averaging. 
However, the combination of these design decisions can hamper the encoding of temporal variation of signals on a larger scale than the size of the convolution filters. To address this, the authors of MiniROCKET propose to use \textit{dilated convolutions}. A dilated convolution virtually increases a filter kernel by adding sequences of zeros in between the values of the original filter kernel \cite{Yu2016} (e.g. [-1 2 1] becomes [-1 0 2 0 1] or [-1 0 0 2 0 0 1] and so on).

\begin{figure}[t]
\centerline{\includegraphics[width=\linewidth]{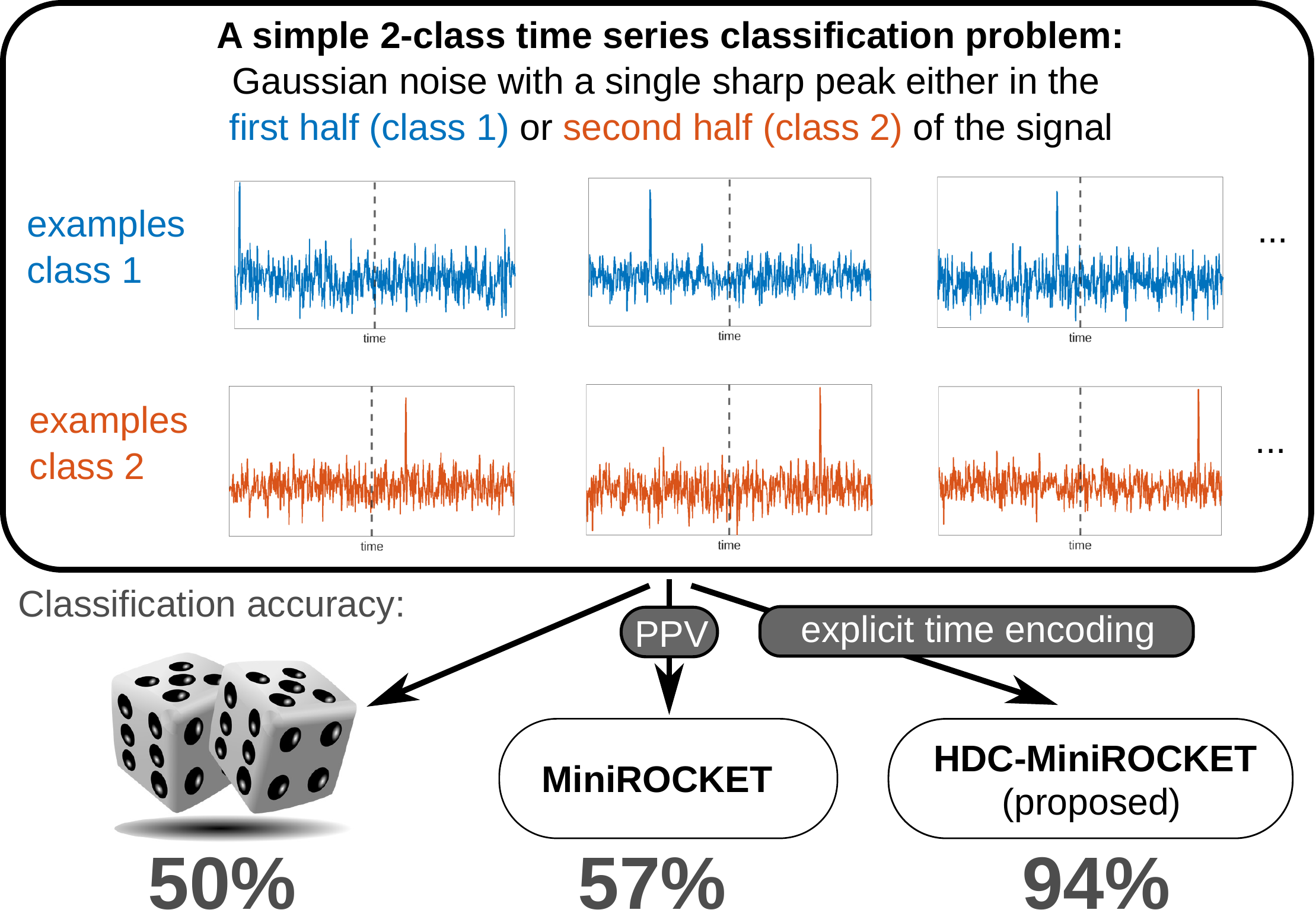}}
\caption{MiniROCKET is a fast state-of-the-art approach for time series classification. However, it is easy to create simple datasets where its performance is similar to random guessing. The proposed HDC-MiniROCKET uses explicit time encoding to prevent this failure at the same computational costs.}
\vspace{-0.5cm}
\label{fig:synth_explain}
\end{figure}

The first contribution of this paper is to demonstrate that although the dilatated convolutions of MiniROCKET perform well on a series of standard benchmark datasets like UCR \cite{Dau2019}, 
it is easy to create datasets where classification based on MiniROCKET is not much better than random guessing.
An example is illustrated in Fig.~\ref{fig:synth_explain}. 
There, the task is to distinguish two different classes of time series signals. Each consists of Gaussian noise and a single sharp peak either in the first half of the signal (for the first class) or in the second half of the signal (for each sample from the second class). 
Since this is a 2-class problem, random guessing of the class of a query signal can achieve 50\% accuracy. Surprisingly, the MiniROCKET implementation achieves an only slightly better accuracy of 57\% after training 
(detail on this experiment will be given in Sec.~\ref{subsec:synth_data}).
This is due to a combination of two problems: 
(1) The location of the sharp peaks cannot be well captured by dilated convolutions with high dilation values due to their large gaps. 
(2) Although responses of filters with small or no dilation can represent the peak, the averaging implemented by PPV removes the temporal information on the global scale.  

Sec.~\ref{sec:approach} will present a novel approach, HDC-MiniROCKET, that addressed this second problem.
It is based on the observation that MiniROCKET's Proportion of Positive Values (the second design decision from above) is a special case of a broader class of accumulation operations known as \textit{bundling} in the context of Hyperdimensional Computing (HDC) and Vector Symbolic Architectures (VSA) \cite{Kanerva09,Plate94Phd,gayler03,Eliasmith07}. 
This novel perspective encourages a straight-forward and computational efficient usage of a second HDC operator, binding, in order to explicitly and flexibly encode temporal information during the accumulation process in MiniROCKET.
The original MiniROCKET then becomes a special case of the proposed more general HDC-MiniROCKET.
As is illustrated in Fig.~\ref{fig:synth_explain} and elaborated in Sec.~\ref{subsec:synth_data}, this explicit time encoding can significantly improve the performance on the above classification problem to 94\% accuracy.
Sec.~\ref{subsec:results_UCR} will demonstrate that the performance also improves on a large collection of standard time series classification benchmark datasets.
The proposed extension can be implemented efficiently as will be discussed in Sec.~\ref{subsec:eff_impl} and evaluated in Sec.~\ref{subsec:results_effort}.

\section{Related Work}

\subsection{Time Series Classification}
\label{subsec:tsc}

The related work of time series classification can basically be divided into two different domains: univariate and multivariate time series. 
For the first case, there is a large number of algorithms in the literature, while the second case is a more general problem formulation with multiple input channels and only recently received increased attention.
There exists are two major survey paper for these problems: \cite{Bagnall2017} for univariate and \cite{Ruiz2021} for multivariate time series classification.
The survey papers used dataset collections for benchmarking: 1) the University of California, Riverside (UCR) time series classification and clustering repository \cite{Dau2019} for univariate time series, and 2) the University of East Anglia (UEA) repository \cite{Bagnall2018} for multivariate time series. 

Bagnall et. all \cite{Bagnall2017} compared a large variety of algorithms from different feature encoding categories like shapelets, dictionary-based or combinations of these.
According to this survey, the most accurate algorithms for univariate time series classification are BOSS (Bag-of-SFA-Symbols) \cite{Schafer2015} (with the more accurate but higher memory consumption extension WEASEL \cite{Schafer2017}), Shapelet Transform \cite{Bostrom2015} and COTE (superseded by HIVE-COTE \cite{Lines2018}), which is an ensemble and uses multiple other classifiers.
BOSS uses SFA (symbolic fourier approximation) \cite{Schafer2012}, which is an accurate Bag of Patterns (BoP) method and is based on the Fourier transformation of the signal.
WEASEL extended the BOSS approach by an additional word extraction from SFA with histogram calculation. 

Besides the survey paper by \cite{Bagnall2017}, there are other recent accurate classification methods such as ensemble classifiers like TS-CHIEF \cite{Shifaz2020} and HIVE-COTE 2.0 \cite{Middlehurst2021}, the model-based neural network InceptionTime \cite{IsmailFawaz2020} (based on the Inseption architecture), and ROCKET \cite{Dempster2020} with its updated version MiniROCKET \cite{Dempster2021} as a simple convolution-based method. 
MiniROCKET is characterized by a very good trade-off between classification accuracy and model complexity and is currently state of the art for univariate time series.

\cite{Ruiz2021} provides an overview of approaches for multivariate time series classification.
CIF \cite{Middlehurst2020} is an improvement of HIVE-COTE \cite{Lines2018} and shows very good accuracies on the mutlivariate benchmark, but both are extremely slow and have high memory consumption (according to \cite{Ruiz2021}: CIF is accurate but takes 6 days for all 30 UEA datasets and HIVE-COTE would take over a year).
WEASEL plus MUSE \cite{Schafer2017b} is an extension of WEASEL \cite{Schafer2017} for the multivariate domain and provides good results but also lags on computation efficiency. 
ROCKET was extended to multivariate classification in \cite{Ruiz2021} and nominated as the clear winner of the multivariate classification benchmark - it achieves high accuracy and is by far the fastest classifier.
Its superseded version MiniROCKET~\cite{Dempster2021} has a recent implementation for multivariate time series and has almost the same accuracy as ROCKET by even lower computational costs.  

Therefore, MiniROCKET with its overall good performance is a good starting point for further improvements in both univariate and multivariate time series classification. 

\subsection{ROCKET and MiniROCKET}
\label{subsec:minirocket}

As said before, MiniROCKET \cite{Dempster2021} is a variant of the earlier ROCKET \cite{Dempster2020} algorithm.
Based on the great success of convolutional neural networks, both variants build upon convolutions with multiple kernels.
However, learning the kernels is difficult if the dataset is too small, so \cite{Dempster2020} uses a fixed set of predefined kernels.
While ROCKET uses randomly selected kernels with a high variety on length, weights, biases and dilations, MiniROCKET is more deterministic. 
It uses predefined kernels based on empirical observations of the behavior of ROCKET. 
Furthermore, instead of using two global pooling operation in ROCKET (max-pooling and proportion of positive values, PPV), MiniROCKET uses just one pooling value -- the PPV. 
This leads to vectors that are only half as large (about 10,000 instead of 20,000 dimensions).
MiniROCKET is up to 75 times faster than ROCKET on large datasets.
To classify feature vectors, both ROCKET and MiniROCKET use a simple ridge regression. 
Although MiniROCKET uses less feature dimensions, the accuracy of both approaches is almost identical.

\subsection{Hyperdimensional Computing (HDC)}
Hyperdimensional computing (also known as Vector Symbolic Architectures, VSA) is an established approach to solve computational problems using large numerical vectors (hypervectors) and well-defined mathematical operations.
Basic literature with theoretical background and details on implementations of HDC are \cite{Kanerva09,Plate94Phd,gayler03,Eliasmith07}; further general comparisons and overviews can be found in \cite{Kleyko2021a, Kleyko2021b, FradyKS18,Schlegel2020}.
HDC has been applied in various fields including addressing catastrophic forgetting in deep neural networks \cite{CheungTCAO19}, image feature aggregation \cite{Neubert2021b}, semantic image retrieval \cite{Neubert2021a},  medical diagnosis \cite{Widdows15}, robotics \cite{Neubert19}, fault detection \cite{Kleyko15a}, analogy mapping \cite{Rachkovskij12}, reinforcement learning \cite{Kleyko15}, long-short term memory \cite{Danihelka16}, text classification \cite{Kleyko18}, and synthesis of finite state automata \cite{Osipov17}.
Moreover, hypervectors are also intermediate representations in most artificial neural networks. 
Therefore, a combination with HDC can be straightforward. 
Related to time series, for example, \cite{Neubert19} used HDC in combination with deep-learned descriptors for temporal sequence encoding for image-based localization. 
A combination of HDC and neural networks for multivariate time series classification of driving styles was demonstrated in \cite{Schlegel2021b}. 
There, the HDC approach was used to first encode the sensory values and to then combine the temporal and spatial context in a single representation. 
This led to faster learning and better classification accuracy compared to standard LSTM neural networks.

\section{A Primer: The algorithmic steps of MiniROCKET \cite{Dempster2021}}
Input is a time series signal $x \in \mathbb{R}^T$ where $T$ is the length of the time series.
MiniROCKET can compute a $D=9,996$ dimensional output vector $y$ that describes this signal.
For application to time series classification, given training data of time signals with known class, a classifier (e.g. a simple ridge regression or a neural network) can be trained  on these descriptor vectors and then be used to classify descriptors of new time series signals.

The first step in MiniROCKET is a dilated convolution of the input signal $x$ with kernels $W_{k,d}$:
\begin{equation}
 c_{k,d} = x * W_{k,d}
 \label{eq:c_kd}
\end{equation}
$d$ is the dilation parameter, $k\in\{1,...,84\}$ refers to 84 pre-defined kernels $W_k$. 
The length and weights of these kernels are based on insights from the first ROCKET method \cite{Dempster2020}:
the MiniROCKET kernels have a length of 9, the weights are restricted to one of two values $\{-1, 2\}$, and there are exactly three weights with value 2. 
To extend the receptive field of the different kernels, each kernel is applied with one or more dilations $d$ selected from an exponential distribution to ensure that exponentially more features are computed with smaller dilations.
For details on the kernels and dilations, please refer to \cite{Dempster2021}.

In a second step, each convolution result $c_{k,d}$ is element-wise compared to one or multiple bias values $B_b$: 
\begin{equation}
 c_{k,d,b} = c_{k,d} > B_b
 \label{eq:c_kdb}
\end{equation}
This is an approximation of the cumulative distribution function of the filter responses of this particular combination of kernel $k$ and dilation $d$.
MiniROCKET computes the bias values based on the quantiles of the convolution responses of a randomly chosen training sample. The number of bias values is such that there are exactly 119 different combinations of dilations $d$ and bias values $B_b$ for each of the 84 kernels $W_i$, resulting in $119 \cdot 84 = 9,996$ different binary vectors $c_{k,d,b} \in\{0,1\}^T$ ($T$ is the length of the input time series $x$). Again, for the details, please refer to \cite{Dempster2021}.

The final step of MiniROCKET is to compute each of the 9,996 dimensions of the output descriptor $y^{PPV}$ as the mean value of one of the $c_{k,d,b}$ vectors (referred to as PPV in \cite{Dempster2021}).
This averaging is where potentially important temporal information is lost and this final step will be different in HDC-MiniROCKET.

\section{Algorithmic approach: HDC-MiniROCKET}
\label{sec:approach}

The proposed HDC-MiniROCKET is a variant of MiniROCKET. Both use the same set of dilated convolutions to encode the input signal. The important difference is that HDC-MiniROCKET uses a Hyperdimensional Computing (HDC) binding operator to bind the convolution results to timestamps before creation of the final output. 
For a general introduction to HDC and explanations, \textit{why} these operators work in this application, please refer to one of \cite{Kanerva09,Plate94Phd,gayler03,Eliasmith07}. 
Here, we will only very briefly introduce some of the concepts and explain their usage for encodings in HDC-MiniROCKET.
Sec.~\ref{subsec:hdc} introduces HDC bundling and formulates MiniROCKET in terms of this operation. 
Sec.~\ref{subsec:hdc_minirocket} introduces the second operator, binding, and uses it in combination with time encodings based on fractional binding in the proposed HDC-MiniROCKET.
Finally, Sec.~\ref{subsec:eff_impl} will present an efficient implementation.

\subsection{A HDC perspective on MiniROCKET}
\label{subsec:hdc}
The general concept of HDC is to systematically process very high-dimensional vectors with carefully designed vector operations. 
One of these operations is \textit{bundling} $\oplus$. Input to the bundling operation are two or more vectors from a vector space $\mathbb{V}$ and the output is a vector of the same size that is similar to each input vector.
Dependent on the underlying vector space $\mathbb{V}$, there are different Vector Symbolic Architectures (VSAs) that implement this operation differently. For example, the multiply-add-permute (MAP) architecture \cite{gayler03}
can operate on bipolar vectors (with values $\{-1,1\}$) and implements bundling with a simple element-wise summation -- in high-dimensional vector spaces, the sum of vectors is similar to each summand~\cite{Plate94Phd}.

The valuable connection between HDC and MiniROCKET is that the 9,996 different vectors $c_{k,d,b} \in\{0,1\}^T$ from eq. \ref{eq:c_kdb} in MiniROCKET also constitute a 9,996-dimensional feature vector $F_t$ for each timestep $t\in \{1,...,T\}$ (think of each vector $c_{k,d,b}$ being a row in a matrix, then these feature vectors $F_t$ are the columns).
The averaging in MiniROCKET's PPV (i.e. the row-wise mean in the above matrix) is then equivalent to bundling of these vectors $F_t$. 
More specifically, if we convert $F_t$ to bipolar by $F^{BP}_t = 1-2F_t$ and use the MAP architecture's bundling operation (i.e. elementwise addition) then the result is proportional to the output of MiniROCKET:
\begin{equation}
 \bigoplus_{t=1}^T F^{BP}_t = \sum_{t=1}^T (1-2F_t) \propto y_{PPV}
\end{equation}

This is just another way to compute output vectors that point in the same directions as the MiniROCKET output -- which preserves cosine similarities between vectors.
But much more importantly, it encourages the integration of a second HDC operation, \textit{binding} $\otimes$, as described in the following section.

\subsection{From MiniROCKET to HDC-MiniROCKET}
\label{subsec:hdc_minirocket}

\begin{figure}[t]
\centerline{\includegraphics[width=0.8\linewidth]{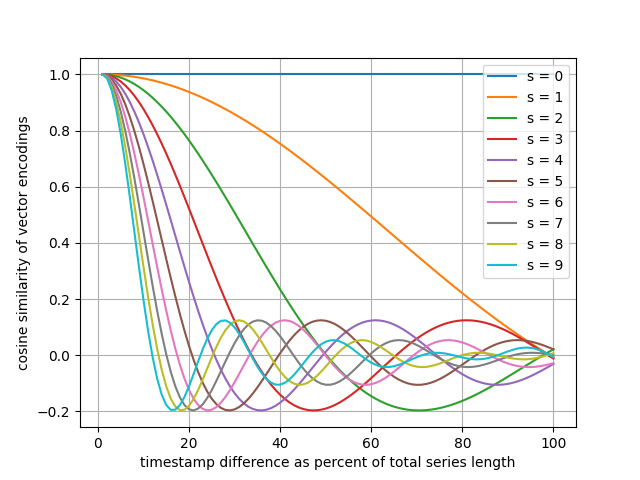}}
\caption{Different graded similarities for timestamp encodings.}
\vspace{-0.5cm}
\label{fig:timestamps_sim}
\end{figure}

HDC-MiniROCKET extends MiniROCKET by using the HDC \textit{binding} $\otimes$ operation to also encode the 
timestamp $t$ of each feature $F^{BP}_t$ in the output vector (without increasing the vector size).

In HDC, the input to the binding $\otimes$ operation are two vectors from the same vector space and the output is a single vector, again of the same size, that is non-similar to each of the input vectors. We will use the property that binding is similarity preserving, i.e.,  $ \forall a,b_1, b_2 \in \mathbb{V}: sim(a \otimes b_1, a \otimes b_2) \approx sim(b_1,b_2)$.
A typical usage of binding is to encode key-value pairs.
In the MAP architecture, binding is implemented by simple element-wise multiplication \cite{gayler03}. 

The output of HDC-MiniROCKET $y^{HDC}$ is computed by:
\begin{equation}
 y^{HDC} = \bigoplus_{t=1}^T (F^{BP}_t \otimes P_t) = \sum_{t=1}^T (1-2F_t) \odot P_t
 \label{eq:y_hdc}
\end{equation}

where $P_t$ is a systematic encoding of the timestamp $t$ in a vector of the same length as $F^{BP}_t$, i.e. 9,996-dimensional.
Before we provide details on the creation of $P_t$, we want to emphasize that all vectors, including $y^{HDC}$, are of the same length and that all operations are simple and efficient local operations. In the MAP architecture, $\oplus$ is simple element-wise addition, $\otimes$ is element-wise multiplication.

The vectors $P_t$ are intended to encode temporal position by making early events dissimilar to later ones. 
For creating such hypervectors, we use fractional binding \cite{Komer2019}.
It uses a predefined real valued hypervector $\mathbf{v}$ of dimensionality $D$ and transfers a scalar value $p$ to a hypervector by applying the following equation: 
\begin{equation}
  \mathbf{v}^{p} := IDFT \left(\left(DFT_{j}\left(\mathbf{v}\right)^{p}\right)_{j=0}^{D-1}\right).
  \label{eq:conv_power}
\end{equation}
DFT and IDFT are the Discrete Fourier Transform and the Inverse Discrete Fourier Transform.
The resulting vector $\mathbf{v}^p$ represents a systematically encoded hypervector of the scalar $p$, which means that the procedure preserved the similarity of scalar values (euclidean distance) within the hyperdimensional vector space (cosine similarity). 

To be able of adjust the temporal similarity of feature vectors, we introduce a parameter $s$, which influences the graded similarity between consecutive timestamps. 
In HDC-MiniROCKET the scalar values $p_t$ at time $t$ in a time series with length $T$ is given by  $p_t = \frac{t \cdot s}{T}$.
The scale factor $s$ adjusts how fast the similarity of timestamp encodings decreases with increasing difference in $t$. 
This is visualized in Fig.~\ref{fig:timestamps_sim}.
If $s$ is equal to 0, all timestamp encodings are the same.
$s=1$ means the similarity decreases for increasing difference of timestamps and only reaches zero for the difference between the first and the last timestamp of the whole time series.
$s=2$ means that the  similarity went to zero after the half of the length, and so on.
$s$ weigths the importance of temporal position -- the higher its value, the more dissimilar become timestamps.

This systematic encoding of timestamps is exploited in HDC-MiniROCKET in Eq.~\ref{eq:y_hdc}. Although binding $\otimes$ and bundling $\oplus$ can be implemented as simple as element-wise multiplication and addition, these operations have partially surprising properties in very high-dimensional vector spaces such as the 9,996-dimensional MiniROCKET features. 
In HDC-MiniROCKET, the underlying mechanism is the similarity preserving property of the HDC-binding operation $\otimes$ from the beginning of this section: the similarity of two feature vectors, each bound to the encoding of their timestamp, gradually decreases with increasing difference of the timestamps. This information is maintained when bundling ($\oplus$) feature vectors to create the output $y^{HDC}$, since the output of bundling is similar to each input. For more details on these underlying mechanisms of hyperdimensional computing, please refer to one of \cite{Schlegel2020,Neubert19,Kleyko2021a}.

As described earlier, the proposed temporal encoding creates a more general variant of MiniROCKET -- if parameter $s$ is set to 0, the cosine similarities based on HDC-MiniROCKET are identical to those of MiniROCKET.
The higher $s$, the more distinctive is the temporal context inside HDC-MiniROCKET's feature calculation.
The selection of the value of $s$ will be discussed and evaluated in Sec.~\ref{subsec:results_scale}.

\subsection{Efficient implementation}
\label{subsec:eff_impl}
A low computational complexity is an important feature of MiniROCKET. 
HDC-MiniROCKET can be implemented with the same (in fact, even slightly \textit{lower}) computational complexity during inference. 

The efficient implementation of HDC-MiniROCKET builds on two simple observations: 
First, $P_t$ can be pre-computed for all $t$ since it only depends on the fixed seed vector $v$ and the scaling parameter $s$, i.e. it is independent of the signal $x$.
Second, we can omit all additions and multiplications when computing $F_t^{BP}\oplus P_t = (1-2F_t)\odot P_t$ in Eq.~\ref{eq:y_hdc}.
The term $(1-2F_t)$ in Eq.~\ref{eq:y_hdc} can be directly computed by replacing the binarization in MiniROCKET from Eq.~\ref{eq:c_kdb}:by: 
\begin{equation}
 c'_{k,d,b} = 
 \begin{cases} 
 1 & if \ c_{k,d} > B_b \\
 -1 & if \ c_{k,d} < B_b
 \end{cases}
 \label{eq:c_kdb_simply}
\end{equation}

\noindent
This can also directly integrate the mutiplication with $P_t$:
\begin{equation}
 c''_{k,d,b} = 
 \begin{cases} 
 P_{t,i} & if \ c_{k,d} > B_b \\
 -P_{t,i} & if \ c_{k,d} < B_b
 \end{cases}
\label{eq:simple_binding}
\end{equation}
Here, $P_{t,i}$ is the i-th dimension of the vector time encoding of time step $t$.
Finally, the i-th dimension of the output of HDC-MiniROCKET is then simply:
\begin{equation}
    y^{HDC}_i = \sum_{t=1}^T c''_{k,d,b}
\label{eq:y_hdc_eff}
\end{equation}

\noindent
In summary, the computational steps of HDC-MiniROCKET are: 
\begin{enumerate}
\item Compute the filter responses $c_{k,d}$ for each kernel-dilation combination (Eq.~\ref{eq:c_kd}). 
\item Compare each of them to a set of bias values $B_b$ to select either a positive or negative precomputed time encoding $P_{t,i}$ (Eq.~\ref{eq:simple_binding}).
\item Compute the sum over all time steps (Eq.~\ref{eq:y_hdc_eff}).
\end{enumerate}

Other than the precomputation of time encodings $P_t$, there are no additional computations compared to MiniROCKET. 
The PPV computation of the original MiniROCKET even requires an additional division per output dimension; however, if cosine similarities are used, then this division could also be omitted in the original MiniROCKET.

\section{Experimental Evaluation}
We evaluate the performance of the proposed HDC-MiniROCKET in two different setups: (1) the simple synthetic dataset classification task from the introduction and Fig.~\ref{fig:synth_explain}, and (2) the 128 datasets from the UCR benchmark for univariate time series classification.
Finally, we will analyze the computational efficiency of our implementation. 

We use a Python implementation based on the code of MiniROCKET from the Python library sktime\footnote{sktime, GitHub, \url{https://github.com/alan-turing-institute/sktime}}, which contains the original code\footnote{minirocket, GitHub, \url{https://github.com/angus924/minirocket}} of \cite{Dempster2021}. 
For classification, we also use the same ridge regression classifier as the original MiniROCKET~\cite{Dempster2021}.

\subsection{Synthetic Dataset}
\label{subsec:synth_data}

As described before, the global pooling by PPV in MiniROCKET can neglect potentially important temporal information if it is not well captured by the dilated convolutions.
To demonstrate possible problems of such behavior, we constructed a simple synthetic dataset, which consists of two classes with high temporal dependency. 
Example signals for both classes are shown in Fig.~\ref{fig:synth_explain}.
Each signal consists of a single sharp peak either in the first (class 1) or second (class 2) half of the signal and additional Gaussian noise ($\mu=0$, $\sigma=1$).
The peak is created using an approximate delta function, which is based on the normal distribution function: 
\begin{equation}
\delta(t) = \frac{1}{\sqrt{2 \pi a}} \cdot e^{-\frac{t^{2}}{2 a}}
\end{equation}
The parameter $a$ influences the shape of the peak. We use $a=0.03$ which creates a high and sharp peak that resembles a Dirac impulse.

The length of the time series is 500 time steps. For each time step, we create one sample in the dataset with the peak centered at this position. Accordingly, the dataset $D$ consists of 500 samples, 250 samples of each class.  

Classification with a random split of 80\% training and 20\% test data leads to the accuracies in table \ref{tab:results_synthetic} second column (``standard case``). 
It can be seen that MiniROCKET can correctly classify only 65\% of the test samples, while the HDC version with $s=1$ can achieve 97\%. 
An even more drastic result is seen if we select a particularly challenging subset from the samples; i.e., those class-1 and class-2 sample pairs of the original MiniROCKET-transformed signals that have high similarity and can be easily confused. 
The result is a dataset with 250 highly similar data samples that leads to the classification accuracies in the rightmost column of table~\ref{tab:results_synthetic} (again using a 80\%-20\% split). 
In this challenging scenario, MiniROCKET without global temporal encoding is not significantly better than random guessing. 
However, when HDC-based temporal encoding is added to MiniROCKET, it can solve the classification problem with 94.1\% accuracy. 

\begin{table}[t]
\caption{Results on the synthetic dataset of random guessing and MiniROCKET without and with temporal encoding}
\vspace{-0.4cm}
\begin{center}
\begin{tabular}{|l|c|c|}
\hline
\textbf{Method} & \textbf{Accuracy [\%]} & \textbf{Accuracy [\%]} \\
& \textbf{standard case} & \textbf{challenging case} \\
\hline
random guess & 50.0 & 50.0 \\
\hline
MiniROCKET & 65.0 & 56.8	\\
\hline
HDC-MiniROCKET (s=1) & 97.0 & 94.1 \\
\hline
\end{tabular}
\label{tab:results_synthetic}
\end{center}
\vspace{-0.5cm}
\end{table}

\begin{figure}[t]
\centerline{\includegraphics[width=\linewidth]{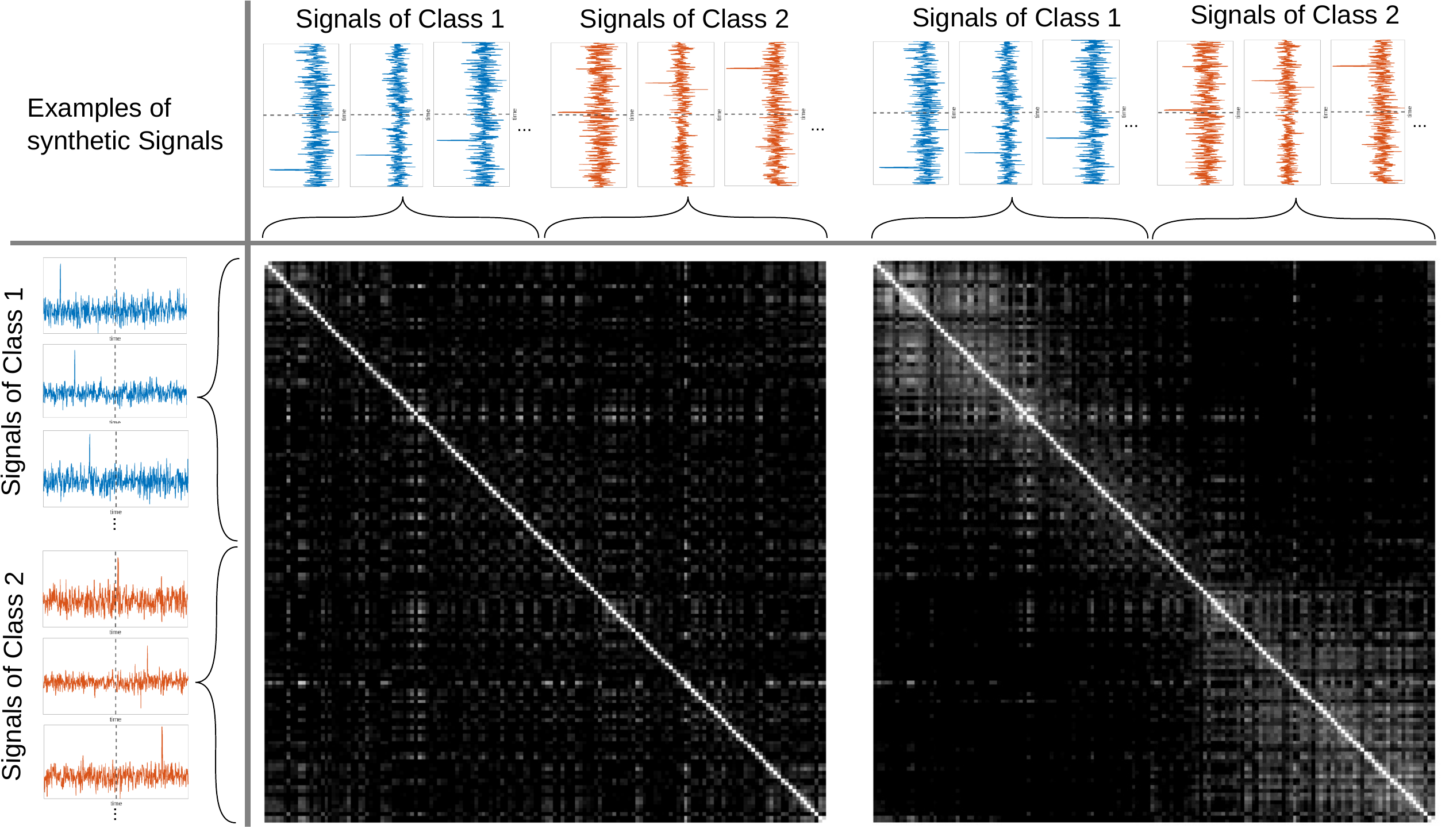}}
\vspace{-0.1cm}
\caption{Similarity matrices on the synthetic dataset. \textit{(left)} Original MiniROCKET \textit{(right)} HDC-MiniROCKET with explicit temporal encoding. Bright pixels indicate high and dark low similarity. On top and left to the matrices are overlayed examples of the synthetic signals colored in red for class 1 and blue for class 2.}
\vspace{-0.5cm}
\label{fig:synthetic_dataset}
\end{figure}

\begin{figure*}[h]
\centerline{\includegraphics[width=\linewidth]{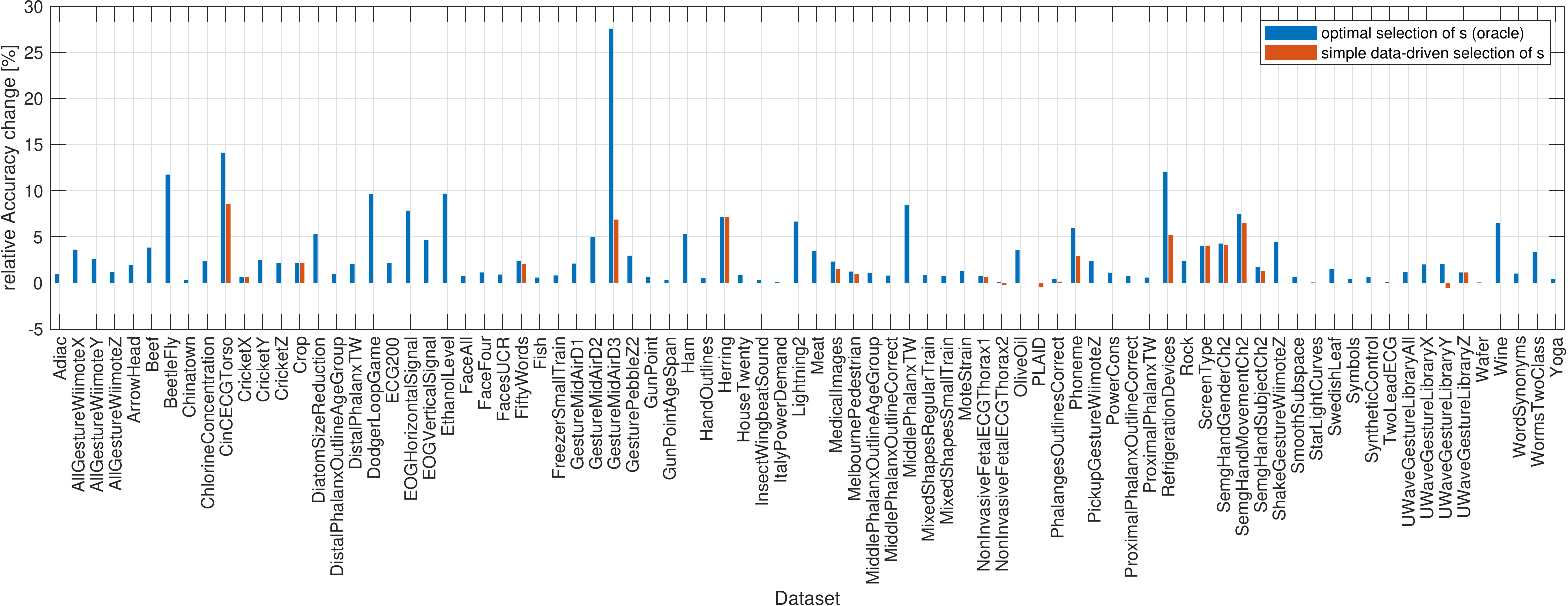}} 
\vspace{-0.2cm}
\caption{Relative accuracy improvements for selecting the optimal $s$ and automatically calculated $s$ while training. Only datasets with a change unequal zero are shown.}
\vspace{-0.5cm}
\label{fig:ucr_results_auto_s}
\end{figure*}

To better illustrate the effect of the time encoding, Fig.~\ref{fig:synthetic_dataset} shows the two similarity matrices of the original encoded MiniROCKET descriptors (left) and the HDC-MiniROCKET descriptors (right). 
The i-th row as well as the i-th column both correspond to the signal from the dataset with the peak at time step i. The matrix entries are the cosine similarities.
The blue and red signals exemplify the shape of samples of classes 1 and 2. 
The first half of the rows and columns of the similarity matrices refer to class 1 and the second half to class 2. 
In the left similarity matrix, it can be seen that for the original MiniROCKET without explicit temporal coding, class 1 signals (red) sometimes resemble class 2 signals (blue) and vice versa. 
In contrast, the right similarity matrix separates the two different classes by explicit temporal coding - class 1 signals are less likely to be similar to class 2 signals. 

\subsection{UCR benchmark}
\label{subsec:results_UCR}

To evaluate the practical benefit of the proposed HDC-MiniROCKET, we show results on the 128 datasets from the UCR benchmark \cite{Dau2019}.
This standard benchmark has previously been used to compare different time series classification algorithms and contains datasets from a broad range of domains \cite{Dau2019}.
Since MiniROCKET showed excellent performance (especially computational efficiency) on this dataset collection compared to other state-of-the-art approaches like \cite{Shifaz2020, Middlehurst2021}, we only compare against MiniROCKET in our experiments.

\begin{figure}[t]
\centerline{\includegraphics[width=0.7\linewidth]{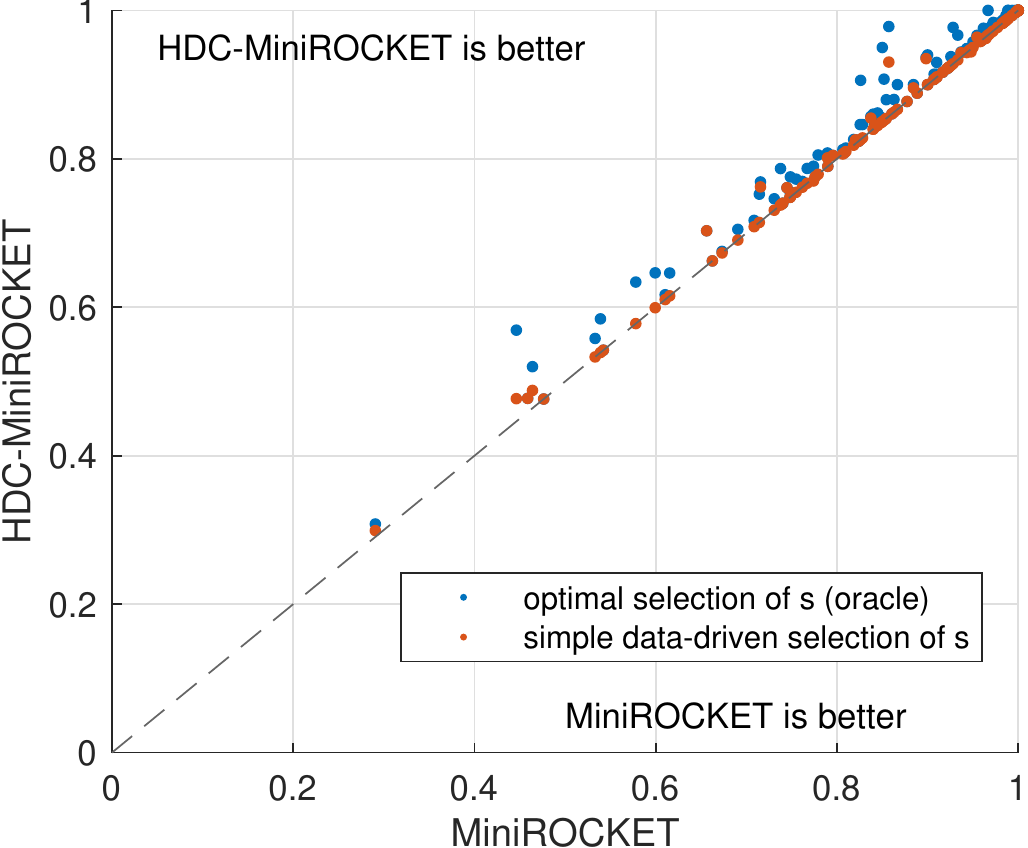}} 
\vspace{-0.1cm}
\caption{Classification accuracies on the 128 UCR datasets when using MiniROCKET or HDC-MiniROCKET. There are two points for each dataset. Blue color is with an oracle that  selects the best scale parameter, orange color is with the simple cross validation for parameter selection. For points in the top-left triangle, HDC-MiniROCKET is better.} 
\vspace{-0.5cm}
\label{fig:ucr_pairwise}
\end{figure}

Fig.~\ref{fig:ucr_pairwise} shows the classification accuracies on all UCR datasets for MiniROCKET in comparison to HDC-MiniROCKET.
The blue colored dots show the performance of HDC-MiniROCKET with an oracle that selects the optimal scale value $s$ for each dataset from the set $\{0,1,2,...,6\}$.
It is important to keep in mind that dependent on the underlying task and dataset, explicit time encoding can be intended and helpful or also unintended and harmful.
This oracle and a simple data-driven approach to select the scale $s$ are discussed in the next Sec.~\ref{subsec:results_scale}.
The evaluation with oracle is intended to demonstrate the potential of the explicit time encoding in HDC-MiniROCKET. 
Fig.~\ref{fig:ucr_results_auto_s} shows the relative change in accuracy for the individual URC datasets. 
For 81 out of 128 datasets, there is an improvement by the explicit time encoding. Although the improvement is sometimes rather small (the average improvement across these 81 datasets is 3.1\%), it also reaches more than 27\% for one dataset. For those datasets without improvement, the performance is the same as for MiniROCKET (which is the special case of $s=0$) -- if we are able to select the optimal scale.

\subsection{Selecting the scale $s$}
\label{subsec:results_scale}

\begin{table*}[t]
\caption{Results for different similarity parameters $s$ on UCR benchmark.}
\vspace{-0.4cm}
\begin{center}
\begin{tabular}{|c|c|c|c|c|c|c|c|c|c|c|c|}
\hline
\textbf{Param. $s$} & 0 & 1 & 2 & 3 & 4 & 5 & 6 & optimal s & automatic selected s\\
\hline
\textbf{Mean} & 0.8540&	0.8537	&0.8500&	0.8450&	0.8407	&0.8366	&0.8336 &0.8681 &0.8569\\
\cline{1-10}
\textbf{Worst Case} & 0.2906	&0.3080	&0.3054	&0.2788	&0.1827	&0.0865	&0.0769 &0.3080 &0.2991\\
\cline{1-10}
\textbf{Best Case} &1.0	&1.0&	1.0	&1.0	&1.0	&1.0&	1.0&	  1.0	&1.0\\
\hline
\end{tabular}
\label{tab:results_ucr_uea}
\end{center}
\vspace{-0.5cm}
\end{table*}

The evaluation from the previous section used an oracle to select the scale $s$.
Since this time scale parameter is physically grounded, we assume that in a practical application, there will be expert knowledge that guides the selection of this parameter. 
When chosing the scale $s$, it is important to keep two properties in mind:

(1) There is no single value of the scale parameter that works for all datasets. 
This is supported by table~\ref{tab:results_ucr_uea} that shows mean, worst-case and best-case performances across all 128 UCR datasets for each scale parameter. No scale choice is considerably better than all others. These average statistics show that the average benefit of scale selection across all datasets is rather small -- however, as illustrated in Fig.~\ref{fig:ucr_results_auto_s}, the benefit for individual datasets can be quite high.

\begin{figure}[t]
\centerline{\includegraphics[width=\linewidth]{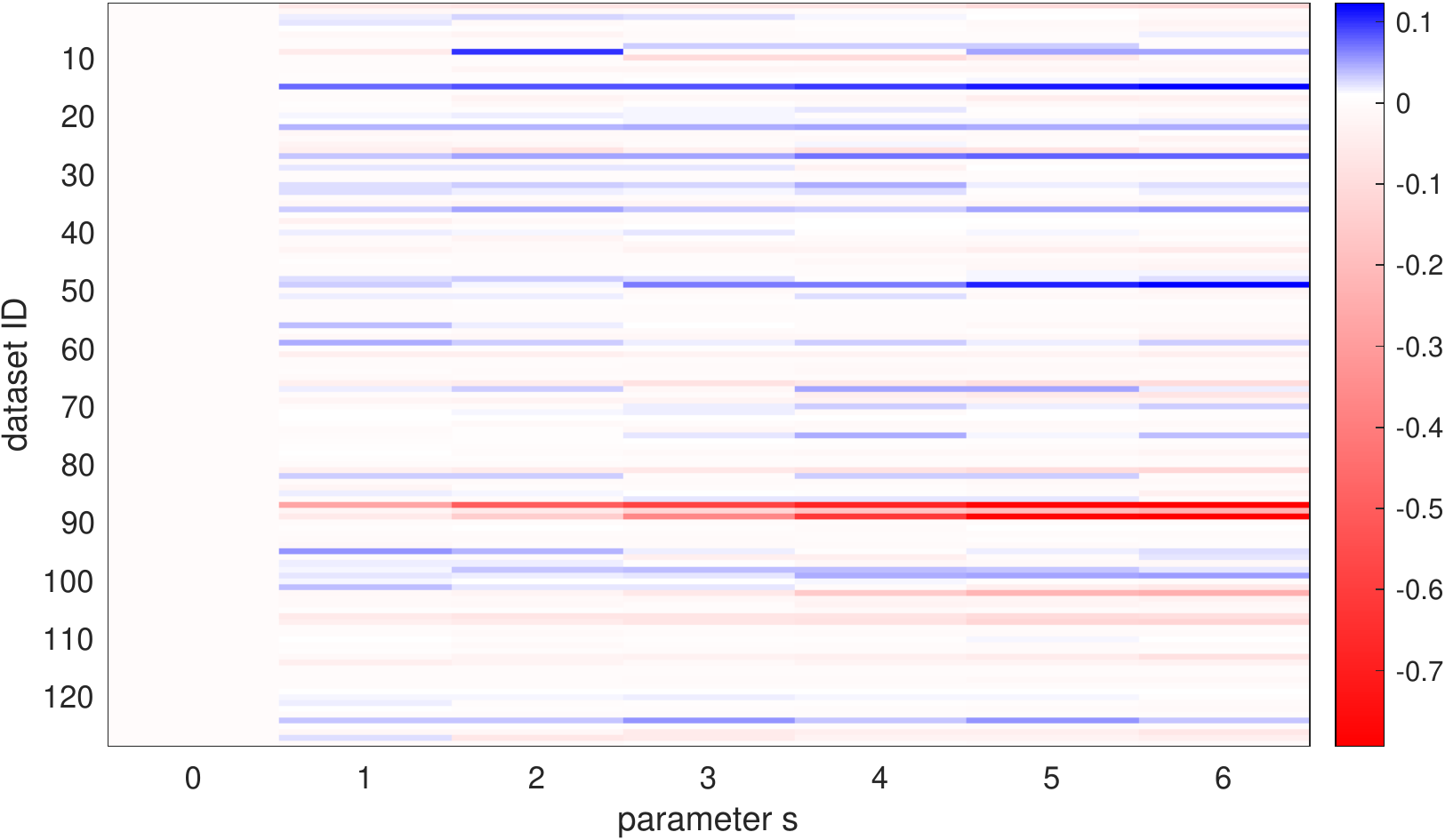}}
\caption{Differences in absolute accuracies between HDC-MiniROCKET and MiniROCKET for different values of $s$.}
\vspace{-0.3cm}
\label{fig:ucr_results}
\end{figure}

(2) The influence of $s$ is quite smooth. Fig.~\ref{fig:ucr_results} shows the influence of different values of $s$ on the performance on individual datasets. Small changes of $s$ create rather small changes of the performance. Typically, there is a single local maximum (which can be one of the extrem values of the evaluated range of $s$).

Although we think that expert knowledge about the task at hand is very valuable to decide whether explicit time encoding is helpful and for selection of $s$, we also propose a purely data-driven approach to select $s$: Since the classifier is trained in a supervised fashion, we can assume a set of labeled training samples. For automatic scale selection, we can simple apply a cross-validation by further splitting the training set into training and validation splits. We use a 10-fold cross-validation to select $s$ from the same possible values as in the oracle. For each fold, we train the classifier on the train split and evaluate on the corresponding validation split (which is, of course, also taken from the UCR train split). We select the value $s$ that performs best in the highest number of splits. Ties are broken in favour of smaller values.

Fig.~\ref{fig:ucr_pairwise} and \ref{fig:ucr_results_auto_s} also provide results for this simple automatic scale selection. 
Although the performance is considerably below that of the oracle, the results are again better than those of the original MiniROCKET.
In particular, since MiniROCKET is a special case with $s=0$, HDC-MiniROCKET with oracle cannot be worse than MiniROCKET. 
Very importantly, even without any knowledge about the tasks, the automatic scale selection is in worst-case only 0.5\% below the performance of MiniROCKET.

\begin{figure}[t]
\centerline{\includegraphics[width=0.7\linewidth]{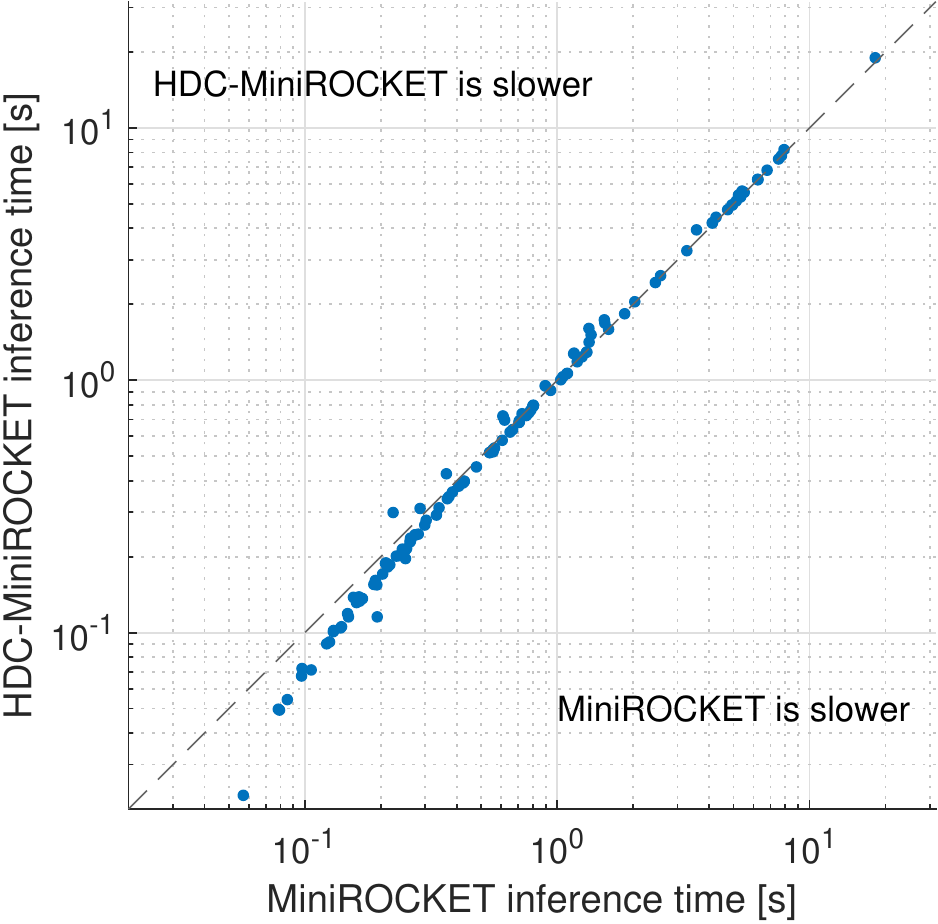}}
\caption{Pairwise comparison of inference time of MiniROCKET and HDC-MiniROCKET.}
\vspace{-0.3cm}
\label{fig:ucr_time_results}
\end{figure}

\subsection{Computational effort}
\label{subsec:results_effort}

As discussed before, apart from the precomputation of time encodings (which only has to be done once), the computational effort for MiniROCKET and HDC-MiniROCKET is almost identical.
Fig.~\ref{fig:ucr_time_results} shows the pairwise comparison of the inference times of both approaches (computed on a standard desktop computer with Intel i7-5775C CPU). 
It can be seen that HDC-MiniROCKET is occasionally slower for some datasets and is consistently marginally faster for fast data sets. 
Presumably this is due to the difference in the calculation of PPV in MiniROCKET that averages values, while HDC-MiniROCKET only accumulates them. 
On average, both algorithms require about 1.57s per dataset.

\section{Conclusion}

We proposed to extend the MiniROCKET algorithm for time series classification by incorporating an explicit temporal context through hyperdimensional computing. Very importantly, the explicit time encoding with HDC does not affect computational costs during inference.
HDC-MiniROCKET is a generalization of MiniROCKET where a parameter $s$ can be used to adjust the importance of time encoding in the descriptor of the time series. MiniROCKET becomes the special variant with $s=0$. 

Experiments on a classification task on a synthetic dataset with high temporal dependences demonstrated that such explicit temporal coding can prevent MiniROCKET from failure. 
An evaluation on the 128 UCR datasets demonstrated the potential of the proposed approach on a standard time series classification benchmark.
It is very important to keep in mind that not all tasks and datasets benefit from explicit temporal encoding (e.g.. we can similarly construct datasets where temporal encoding is harmful).
In HDC-MiniROCKET, this is reflected with the scale parameter $s$.
The experiments demonstrated that the proposed approach with an oracle that selects the best parameter $s$ can achieve classification improvements on about 2/3 of the datasets and up to 27\%. 
In practice, selecting $s$ should incorporate knowledge about the particular task. 
We demonstrated that selecting $s$ using a simple purely data-driven approach increased performance for 17 of the 128 data sets.
However, the presented automatic selection of $s$ is only a simple, prelimary approach -- the very promising results using the oracle demonstrate the general potential of the proposed time encodings and that the automatic scale selection is a promising direction for future research.

\bibliographystyle{IEEEtran}
\bibliography{references}

\end{document}